\documentclass[conference]{IEEEtran}
\IEEEoverridecommandlockouts
\usepackage{cite}
\usepackage{amsmath,amssymb,amsfonts}
\usepackage{algorithmic}
\usepackage{graphicx}
\usepackage{textcomp}
\usepackage{xcolor}
\usepackage{booktabs} 
\usepackage{multirow} 
\usepackage[affil-it]{authblk}
\usepackage{xspace}

\makeatletter
\DeclareRobustCommand\onedot{\futurelet\@let@token\@onedot}
\def\@onedot{\ifx\@let@token.\else.\null\fi\xspace}

\makeatother

\def\BibTeX{{\rm B\kern-.05em{\sc i\kern-.025em b}\kern-.08em
    T\kern-.1667em\lower.7ex\hbox{E}\kern-.125emX}}
\begin{document}

\title{V-CAGE: Vision-Closed-Loop Agentic \\Generation Engine for Robotic Manipulation

}


\author{
    Yaru Liu\textsuperscript{1,*} \quad Ao-bo Wang\textsuperscript{2,*} \quad Nanyang Ye\textsuperscript{3} \\
    \small \textsuperscript{1}University of Cambridge \quad \textsuperscript{2}Wuhan University \quad \textsuperscript{3}Shanghai Jiao Tong University \\
    \small \textsuperscript{*}Equal contribution \\
    \small \texttt{yl962@cam.ac.uk, wangab@whu.edu.cn, ynylincoln@sjtu.edu.cn}
}

\maketitle

\begin{abstract}
Scaling Vision-Language-Action (VLA) models requires massive datasets that are both semantically coherent and physically feasible. However, existing scene generation methods often lack context-awareness, making it difficult to synthesize high-fidelity environments embedded with rich semantic information, frequently resulting in unreachable target positions that cause tasks to fail prematurely. We present V-CAGE (Vision-Closed-loop Agentic Generation Engine), an agentic framework for autonomous robotic data synthesis. Unlike traditional scripted pipelines, V-CAGE operates as an embodied agentic system, leveraging foundation models to bridge high-level semantic reasoning with low-level physical interaction. Specifically, we introduce Inpainting-Guided Scene Construction to systematically arrange context-aware layouts, ensuring that the generated scenes are both semantically structured and kinematically reachable. To ensure trajectory correctness, we integrate functional metadata with a Vision-Language Model based closed-loop verification mechanism, acting as a visual critic to rigorously filter out silent failures and sever the error propagation chain. Finally, to overcome the storage bottleneck of massive video datasets, we implement a perceptually-driven compression algorithm that achieves over 90\% filesize reduction without compromising downstream VLA training efficacy. By centralizing semantic layout planning and visual self-verification, V-CAGE automates the end-to-end pipeline, enabling the highly scalable synthesis of diverse, high-quality robotic manipulation datasets.
\end{abstract}


\section{Introduction}
In recent years, VLA models have demonstrated breakthrough potential in the field of embodied AI and robotic manipulation \cite{chi2025diffusion}. However, the generalization and emergent capabilities of these high-capacity models heavily rely on massive, high-quality expert trajectory data. Collecting multi-step interaction data in the real world is not only prohibitively expensive but also struggles to adequately cover long-tail scenarios \cite{o2024open,khazatsky2024droid}. Consequently, leveraging high-fidelity physics simulation for large-scale, automated data collection has emerged as an a highly promising approach to overcome the ``data starvation'' bottleneck \cite{gensim,wang2024robogen}. 

With the evolution of simulation technologies, virtual data synthesis has made significant strides \cite{robotwin2, robotwin, xu2026sage, wang2025tabletopgen}, particularly in foundational tasks like grasping or simple pick-and-place. Yet, as task objectives extend toward multi-step, long-horizon complex interactions, existing paradigms reveal a critical limitation: they lack agentic reasoning. Traditional simulation pipelines operate as "blind" scripts, following pre-defined paths without the cognitive flexibility to handle environmental nuances or execution failures. Breaking through these limitations requires a shift from scripted automation to agentic simulation synthesis—a paradigm where the generation system itself acts as an embodied entity capable of task-oriented scene configuration, semantic reasoning, and closed-loop self-verification \cite{gensim,wang2024robogen}.

Despite extensive exploration in automated scene synthesis and task planning, existing generative algorithms for robotic manipulation tasks exhibit notable limitations in both theoretical frameworks and practical execution. Existing studies commonly adopt random distribution strategies for object placement \cite{robotwin, robotwin2}. However, this context-agnostic generation mechanism is highly prone to severe ``geometric conflicts'', such as object clipping, mutual occlusion, or unreachable target positions, when constructing crowded scenes to support extended tasks. These conflicts fundamentally strip the simulation environment of its semantic rationality. Furthermore, existing scene generation methods are largely confined to visual and collision properties, lacking true action execution feasibility. The generated scenes lack fine-grained annotations of metadata, such as object functional points, grasping poses, and relational semantics, rendering them incapable of supporting large-scale, automated training data synthesis \cite{wang2025tabletopgen}. Crucially, many current simulation task synthesis studies operate on a ``generate-then-execute'' open-loop paradigm. 
Although this approach can barely manage short-horizon tasks, it degrades significantly when synthesizing long-horizon tasks. In the absence of closed-loop visual validation, any subtle, non-error-triggering implicit failure (e.g., a ``missed grasp'') leads to severe error accumulation, rendering the subsequent sequence of actions entirely invalid.


To address these challenges, we introduce V-CAGE (Vision-Closed-loop Agentic Generation Engine), a scalable framework that centralizes perception and reasoning within the data synthesis loop. Our core contribution is the shift toward an agentic architecture that ensures the robust correctness of long-horizon trajectories.
The main contributions of this work are as follows:
\begin{itemize}
    \item We propose Inpainting-Guided Scene Construction (IGSC), a algorithm that automates task-driven object selection and placement, significantly reducing geometric conflicts to ensure physical validity.
    \item We bridge the gap between high-level command and low-level action by integrating functional metadata, enabling an AI agent to autonomously search for and execute feasible sub-tasks by mapping pre-annotated functional points to predefined templates, ensuring the physical and semantic validity of each retrieved interaction.
    \item We implement a VLM-based closed-loop verification mechanism that acts as the agent’s "visual conscience," 
    that effectively severs the error propagation chain and the causal integrity and absolute correctness of synthesized long-horizon trajectories.
    \item We introduce an action-aware perceptually-driven compression algorithm that achieves a filesize reduction exceeding 90\% for simulated datasets.
\end{itemize}
By integrating these components, V-CAGE provides a robust, end-to-end pipeline for the autonomous generation of high-fidelity, causally-consistent robotic datasets at scale.

\section{Related work}
\label{relatedwork}

\subsection{LLM-Generated Robotic Simulation and Data Scaling}
Recent advancements have leveraged Large Language Models (LLMs) to automate the creation of robotic simulation tasks and training data, significantly reducing human engineering effort. Frameworks such as GenSim~\cite{gensim} and RoboGen~\cite{wang2024robogen}  utilize LLMs to synthesize task code and environment specifications, enabling the generation of diverse manipulation skills. Building on this, RoboTwin 2.0~\cite{robotwin2} introduces a large-scale data generator using generative digital twins and extensive domain randomization to improve policy robustness. While these works utilize functional object annotations—such as grasp points—to scale data collection, they primarily operate in an open-loop "generate-and-execute" protocol. Such pipelines often suffer from "silent failures" where the generated code executes without runtime errors but fails to achieve the intended semantic goal, such as an object slipping from a gripper. Our framework addresses this by employing OpenClaw~\cite{openclaw2025} to orchestrate the pipeline and integrating functional point metadata to ensure that synthesized scenes are inherently interactive and ready for automated trajectory collection.

\subsection{Agentic 3D Scene Synthesis}
The field of scene synthesis has evolved from rule-based procedural generation to agentic orchestration. SAGE~\cite{xu2026sage} represents the state-of-the-art in this domain, utilizing a Model Context Protocol (MCP)~\cite{mcp2025} to coordinate layout generators and physics critics. While SAGE achieves exceptional physical stability and visual realism through simulator-in-the-loop validation , its primary objective is the stability of the 3D environment itself. Similarly, Scene Weaver utilizes a self-reflective agent for layout refinement but lacks systematic physical grounding. Our work, driven by the OpenClaw~\cite{openclaw2025} agent, shifts the focus from "scene-level stability" to "task-level semantic correctness." Rather than merely constructing a stable room, our agentic framework automates the end-to-end synthesis of verified manipulation trajectories, ensuring that every object placement serves a functional purpose for the intended embodied task.

\subsection{Scalable Data Generation for Manipulation}
Generating large-scale demonstration data is a core challenge in embodied AI. RoboTwin 2.0 \cite{robotwin2} provides a massive benchmark for bimanual manipulation by utilizing generative digital twins and domain randomization. While RoboTwin \cite{robotwin} and related works like RoboGen \cite{wang2024robogen} and GenSim \cite{gensim} utilize functional object annotations (e.g., grasp points) to scale data collection , they typically operate in an open-loop manner. These systems rely on programmatic success criteria (e.g., coordinate checks), which are prone to "silent failures" where code executes without achieving the actual semantic goal. Our work bridges this gap by integrating functional point-based metadata with an autonomous agent that performs end-to-end data refinement, ensuring that the generated scenes are not only interactive but functionally consistent.

\subsection{Vision-Model-Based Verification}
VLMs have become integral to robotic reasoning and long-horizon planning. Models such as VILA~\cite{vila} and MOKA~\cite{moka} demonstrate strong capabilities in grounding abstract instructions into visual keypoints or affine transformations. However, these approaches typically position the VLM as a planner or controller during inference. In our framework, we repurpose the VLM (e.g., Gemini 3~\cite{gemini3report2025}) as a rigorous automated verifier within the data generation loop. By formulating data validation as a VLM-guided rejection sampling process, we effectively prune trajectories that fail visual success criteria. This closed-loop verification cuts the error propagation chain in complex manipulation sequences, ensuring that the resulting dataset contains only causally valid demonstrations.

\begin{figure*}
  \centering
   \includegraphics[width=\linewidth]{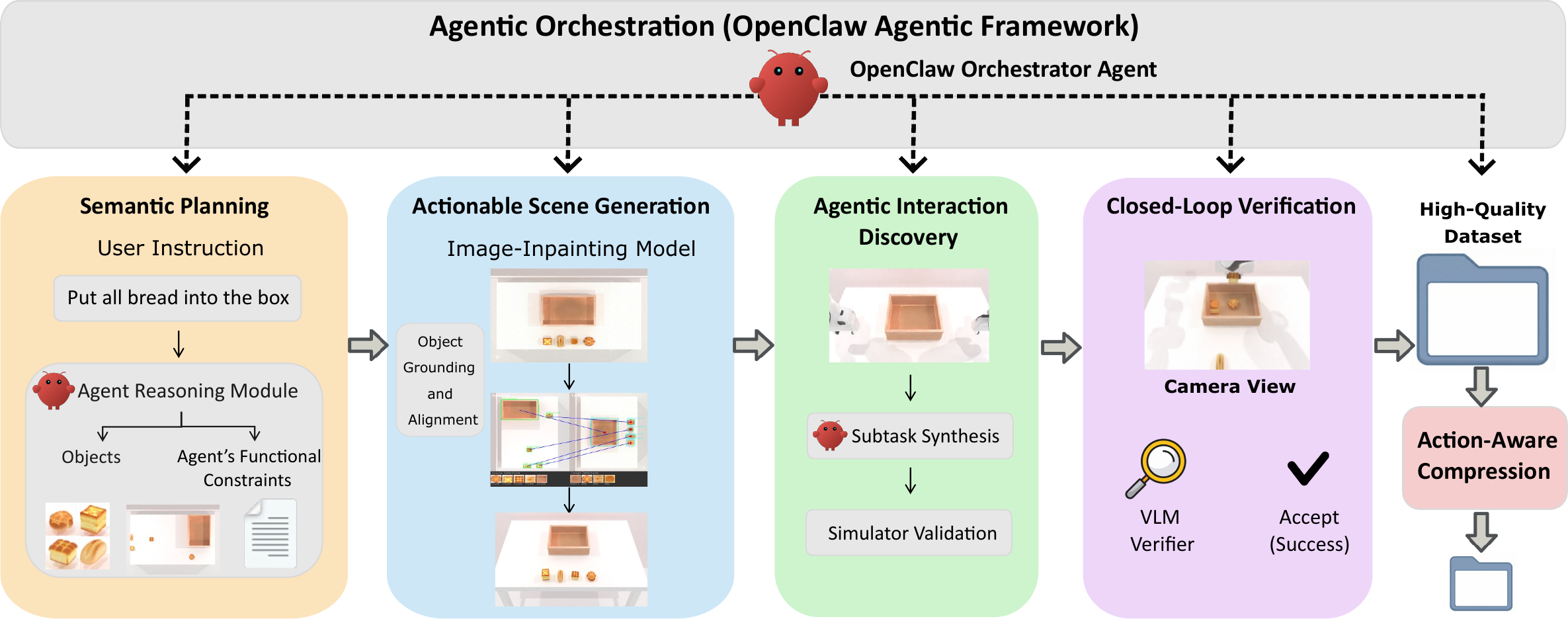}
    \caption{Overview of the V-CAGE framework.This end-to-end engine automates robotic data synthesis by bridging high-level semantic reasoning with IGSC to generate context-aware, physically valid environments across diverse domains. The agentic architecture utilizes VLM-based closed-loop verification to autonomously refine trajectories and sever error propagation chains in long-horizon manipulation tasks. The final output is optimized via action-aware perceptual compression, achieving around 93\% reduction in storage size without compromising visual
fidelity for downstream VLA training.}  
    \label{fig:pipeline}
\end{figure*}

\section{Method}
\label{sec:method}

This section presents the technical architecture of V-CAGE, an end-to-end agentic engine designed to bridge high-level semantic reasoning with low-level physical interaction for autonomous data synthesis. The overall pipeline is illustrated in Fig.~\ref{fig:pipeline}. We build V-CAGE upon the OpenClaw~\cite{openclaw2025} agentic framework, which provides a structured skill-based interface for orchestrating complex simulation workflows. Within this framework, we implement the entire IGSC pipeline as a OpenClaw skill, encapsulating each of its four sequential stages into self-contained, executable scripts that the LLM agent can invoke on demand. This modular design allows the agent to autonomously coordinate scene construction, layout refinement, and trajectory synthesis through a unified skill-calling interface.


\subsection{Inpainting-Guided Scene Construction}
\label{sec:igsc}

Given a natural language task instruction $\mathcal{I}$, the IGSC pipeline aims to automatically construct a simulation scene $\mathcal{S}$ that is both geometrically conflict-free and functionally semantic-aware. The pipeline comprises four sequential stages, each encapsulated as an executable script within the OpenClaw framework for the LLM agent to invoke: (1) LLM-agent-driven asset selection and collision-free initial placement, (2) functional layout planning and inpainting-based refinement, (3) vision-model-based correspondence matching, and (4) collision-free placement optimization.

\subsubsection{LLM-Agent-Driven Asset Selection and Collision-Free Initial Placement}
\label{sec:asset_selection}

Our framework builds upon SAPIEN~\cite{xiang2020sapien} as the physics simulation backend. We maintain a pre-built asset library $\mathcal{A}$ following the RoboTwin~\cite{robotwin} paradigm, wherein each asset is solely annotated with contact points for grasping and functional points for object placement. The geometric properties, such as the 3D bounding box of each object, are calculated dynamically during every algorithmic execution. This annotation strategy ensures that every asset carries sufficient semantic information to support downstream task synthesis without accumulating heavy geometric pre-computations.

Given the task instruction $\mathcal{I}$, an LLM agent is prompted with the full asset catalog to select a semantically relevant subset $\mathcal{A}^* \subset \mathcal{A}$ of $N$ objects, ensuring the chosen items are contextually appropriate for the described task scenario (e.g., selecting bowls, plates, and cutlery for a dining-table task).

Once the asset subset is determined, objects are instantiated in the SAPIEN simulation environment via a collision-free placement procedure. Each object $o_k$ is assigned a position $\mathbf{p}_k = (x_k, y_k, z_k)$ and an orientation quaternion $\mathbf{q}_k$ within a bounded tabletop workspace $\mathcal{W}$. The placement algorithm enforces pairwise non-collision constraints by computing the Axis-Aligned Bounding Box (AABB) for each object---transforming local bounding box corners through the object's rotation matrix and translating to world coordinates---and verifying that no two AABBs intersect. After placement, a top-view image $I_{\text{src}}$ of the scene is rendered, and the world coordinates together with orientation of all objects are recorded in a structured metadata file $\mathcal{M}$.

\subsubsection{Functional Layout Planning and Inpainting-Based Refinement}

\label{sec:inpainting}

While the initial placement ensures geometric validity, it lacks functional semantic structure---objects are randomly scattered without task-meaningful spatial relationships. To inject functional semantics, we employ a two-step process.

First, given the task instruction $\mathcal{I}$ and the scene metadata $\mathcal{M}$, LLM agent generates a natural language layout plan $\mathcal{L}$ that describes the desired spatial relationships aligned with the task's functional requirements (e.g., ``place the bowl on the tray, with the fork beside the plate''). This plan encodes the semantic intent of the task into concrete spatial directives.

Second, the layout description $\mathcal{L}$ together with the source top-view image $I_{\text{src}}$ are fed into an inpainting model (Nano Banana) to produce a target image $I_{\text{tgt}}$. The inpainting model rearranges the visual appearance of objects in the image to reflect the planned layout, while preserving object identity and visual consistency. This step bridges the gap between abstract semantic planning and concrete spatial arrangement without requiring explicit coordinate computation from the LLM agent.

\subsubsection{Vision-Model-Based Correspondence Matching}
\label{sec:matching}

To recover the precise world coordinates of objects in the target layout from the inpainted image $I_{\text{tgt}}$, we perform dense correspondence matching between the source image $I_{\text{src}}$ (where world coordinates are known) and the target image. This stage employs a two-phase pipeline based on vision foundation models.

\paragraph{Object Detection and Feature Extraction.}
We leverage Grounding DINO~\cite{liu2023grounding} as a zero-shot object detector to localize objects in the target image. A text prompt, constructed from the asset names, guides the detector to produce a set of bounding boxes with confidence scores. For each detected region in $I_{\text{tgt}}$ and each known object crop in $I_{\text{src}}$, DINOv2~\cite{oquab2023dinov2} extracts a global feature vector from the CLS token of the cropped region, yielding feature representations $\mathbf{f}_k^{\text{src}}$ and $\mathbf{f}_j^{\text{tgt}}$ for source and target objects, respectively.

\paragraph{Feature Matching and Coordinate Recovery.}
Object correspondences are established via greedy cosine similarity matching: for each source object, we find the best unmatched target detection whose similarity exceeds a threshold $\tau$. For objects that remain unmatched (e.g., due to significant visual changes from inpainting), a semantic name-based fallback mechanism is employed, combining string similarity with feature similarity to recover additional correspondences.

The matched target pixel positions are then converted back to world coordinates via the inverse of the world-to-pixel mapping derived from the source scene's coordinate system. Additionally, per-object rotation estimation is performed by rotating the target crop over a discrete angular grid and computing normalized cross-correlation (NCC) against the source crop, selecting the rotation angle that yields the highest correlation score.

\begin{figure}
  \centering
   \includegraphics[width=\linewidth]{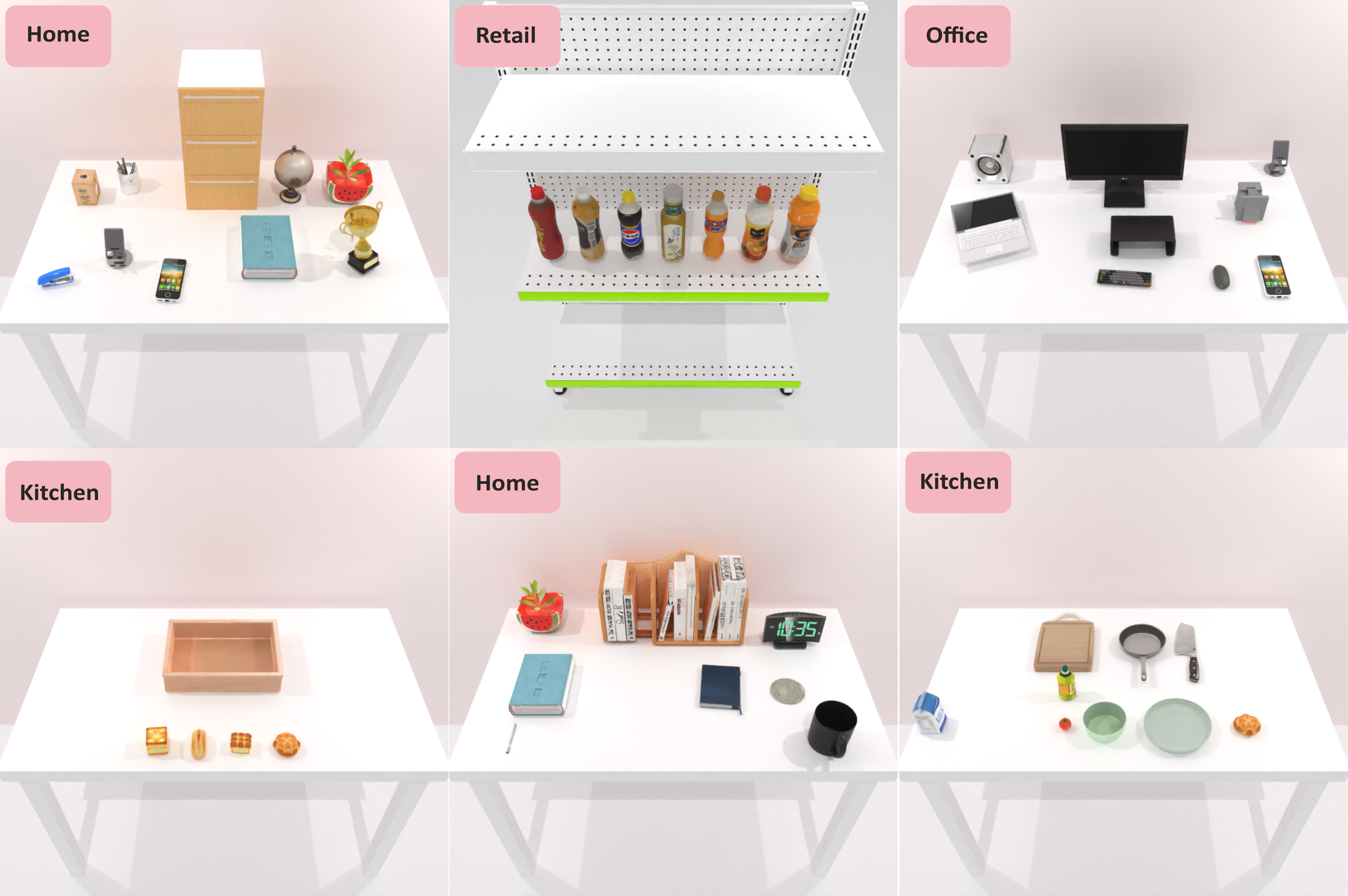}
    \caption{Examples of context-aware scenes generated by V-CAGE, including home, office, retail (supermarket), and kitchen settings.}  
    \label{fig:scenes}
\end{figure}

\subsubsection{Collision-Free Placement Optimization}
\label{sec:refinement}

Due to inherent inaccuracies in feature matching and inpainting, the recovered target coordinates may introduce inter-object collisions. We formulate a constrained optimization problem to refine object positions while preserving the functional layout intent.

Let $\mathbf{x} = [x_1, y_1, \ldots, x_N, y_N]^\top$ denote the concatenated positions of all $N$ objects. We minimize a composite cost function:
\begin{equation}
    \min_{\mathbf{x}} \; \mathcal{J}(\mathbf{x}) = \lambda_c \cdot \mathcal{J}_{\text{coll}}(\mathbf{x}) + \lambda_d \cdot \mathcal{J}_{\text{disp}}(\mathbf{x}) + \lambda_b \cdot \mathcal{J}_{\text{bnd}}(\mathbf{x}),
    \label{eq:total_cost}
\end{equation}
where $\mathcal{J}_{\text{coll}}$ penalizes pairwise AABB penetration and enforces a minimum safety margin $\delta$ between non-container-content object pairs, $\mathcal{J}_{\text{disp}}$ penalizes deviation from the initial target positions with area-based weighting (larger objects are penalized more to preserve the layout backbone), and $\mathcal{J}_{\text{bnd}}$ constrains objects to remain within the workspace boundaries.

Notably, for scenes containing container objects (e.g., trays, bowls), we employ a hierarchical optimization strategy. Objects are first assigned to containers based on collision detection and volumetric analysis. Contained objects are then jointly optimized within container bounds, and finally container groups are optimized as rigid bodies in the global workspace. The optimization is solved using the L-BFGS-B algorithm with multiple random restarts to escape local minima.

The final output of the IGSC pipeline is a refined scene configuration $\mathcal{S}^*$, where each object carries its optimized collision-free position alongside pre-annotated contact points and functional points from the asset library. As shown in Fig.~\ref{fig:scenes}, our IGSC algorithm automatically populates environments with semantically coherent objects, effectively bridging the gap between random placement and structured, task-oriented configurations.

\subsection{Template-Based Sub-Task Search}
\label{sec:subtask_search}

Given the constructed scene $\mathcal{S}^*$ with rich per-object metadata, we efficiently retrieve physically and semantically feasible sub-tasks through a template-based search mechanism. Rather than generating complex robotic scripts from scratch, our approach leverages predefined code templates that encode fundamental robotic manipulation primitives. These templates act as structured code skeletons; LLM agent merely needs to fill in the execution logic by interpreting the task instruction and observing the relevant objects in the scene. To identify valid operations, the algorithm systematically evaluates object properties against template preconditions. Taking the pick-and-place task as an example, the system enumerates object pairs $(o_k, o_l)$ inside the scene. It verifies that the source object $o_k$ possesses valid contact points for grasping, and that the target object $o_l$ features available functional points. An LLM agent filter then semantically verifies whether $o_l$ can serve as a sensible physical receptacle (e.g., a plate or a box) and evaluates the commonsense compatibility of placing $o_k$ onto $o_l$. For each validated pair, a fully executable sub-task script is synthesized. The code generator populates the environment setup with the exact refined coordinates representing the instantiated physical objects, while the LLM agent generates the corresponding action execution and visual success verification logic. As a result, each instantiated sub-task is systematically validated for executability and self-verifying, functioning as a robust atomic building block for the subsequent fully closed-loop validation process.

\subsection{VLM-Based Closed Loop Verification}
\label{sec:vcage}

Even with valid and geometrically consistent sub-tasks generated by our IGSC module, physical execution anomalies---such as gripper slip, slightly misplaced objects, or switches failing to toggle---can still lead to semantic failures during simulation. To filter these ``silent failures,'' we model the data generation pipeline as a rigorous rejection sampling process. We employ a state-of-the-art Vision-Language Model, Gemini 3 \cite{gemini3report2025}, to act as the density estimator and visual critic for valid goal states.

Let $\phi_{\text{VLM}}(I_{\text{img}}, T_i) \in \{0, 1\}$ denote the verification function, where $I_{\text{img}}$ is the post-execution visual observation and $T_i$ is the natural language sub-task description. For a long-horizon trajectory comprising a sequence of $k$ sub-tasks $T_{1:k}$, the entire trajectory is accepted into the final dataset $\mathcal{D}$ if and only if the VLM verifies every step:
\begin{equation}
    \prod_{i=1}^{k} \phi_{\text{VLM}}(I_{\text{img}}^{(i)}, T_i) = 1
\end{equation}
If $\phi_{\text{VLM}}(\cdot) = 0$ evaluates to zero at any intermediate step, the generation episode is immediately aborted, and the scene is reset for re-sampling. This closed-loop filtering mechanism effectively severs the chain of error accumulation in long-horizon execution. It ensures that the final dataset contains only trajectories that are causally and semantically valid, successfully bridging the gap between programmatic code execution and visual reality.

\subsection{Action-Aware Perceptual Compression}
\label{sec:compression_method}
To overcome the "storage wall" while maintaining the data integrity required for high-capacity VLA training, we propose a novel action-aware perceptual compression pipeline. Unlike standard video encoding, our approach prioritizes semantic-critical frames and utilizes a human-perception-driven metric to ensure visual losslessness.

\textbf{Optimized Keyframe Extraction}
Rather than treating all frames equally, our algorithm extracts optimized keyframes $\{I_{k}\}$ based on robotic interaction intensity. We analyze the HDF5 trajectory data to identify gripper state transitions and kinematic velocity peaks.
We locate the midpoints of gripper open/close actions by calculating the median of grouped gripper command changes.
Then we compute the $L_2$ norm of end-effector action vectors $\Delta a$ and extract frames corresponding to the highest motion gradients.By augmenting these with trajectory boundary frames, we condense a sequence of over 60 potential frames into approximately 8 optimized frames per video. These keyframes serve as "stress tests" for the encoder, representing the most informative moments of the manipulation task.

\textbf{Perception-Driven CRF Optimization}
To achieve maximum compression without compromising feature representation, we perform an iterative search for the optimal Constant Rate Factor (CRF) using the HEVC codec.
To capture realistic artifacts, we simulate the end-to-end production environment through an Image $\rightarrow$ Video $\rightarrow$ Image loop. We evaluate quality loss using the ColorVideoVDP metric \cite{mantiuk2023colorvideovdp}. The visual quality is measured in Just-Obvious-Difference (JOD) and we set the threshold as $0.1$ JOD,  $$\text{Loss}_{JOD} = JOD_{\text{reference}} - JOD_{\text{compressed}} < 0.1$$ According to the psychometric mapping in FovVideoVDP \cite{mantiuk2021fovvideovdp}, 0.1 JOD difference implies that the probability of a human observer distinguishing the compressed frame from the original is near 50\%, rendering it visually lossless.
By identifying the highest possible CRF that satisfies the 0.1 JOD constraint across all optimized keyframes, V-CAGE achieves a filesize reduction exceeding 93\%. Experimental validation confirms that this perceptual thresholding preserves the high-frequency spatial details required for downstream VLA training, ensuring that the "learned" physical causality remains identical to that of raw, uncompressed datasets.

\section{Experiments}
\label{sec:experiments}

We empirically evaluate the V-CAGE framework to investigate three primary aspects: (1) the effectiveness of our synthesized datasets in training Vision-Language-Action (VLA) models for complex, long-horizon manipulation; (2) the capability of our video compression strategy to preserve essential visual and semantic features for downstream policy learning; and (3) the contribution of core components—namely, Inpainting-Guided Scene Construction (IGSC) and VLM-based closed-loop verification—to the physical feasibility and purity of the generated data.

\subsection{VLA Policy Training on Long-Horizon Tasks}
\label{subsec:vla_training}

\begin{figure}[ht]
    \centering
    \includegraphics[width=\linewidth]{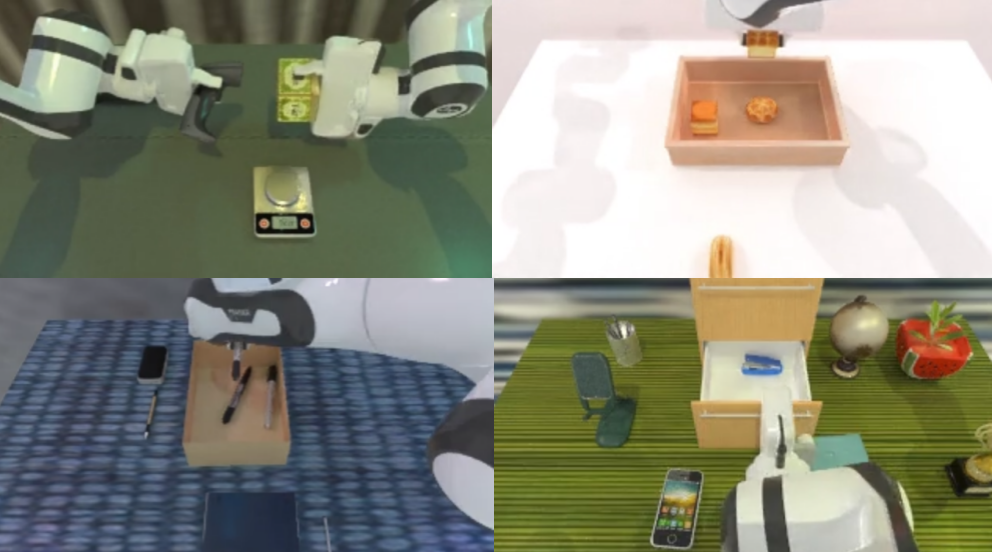}
    \caption{Four complex, long-horizon tasks synthesized by our pipeline.}
    \label{fig:vla_training}
\end{figure}

To demonstrate the utility of our generated data for cutting-edge robot learning, we trained a state-of-the-art $\pi_{0.5}$ VLA model on four complex, long-horizon tasks synthesized by our pipeline. For each task, we generated 100 expert trajectories featuring randomized textures and lighting to encourage robust visual representation. The four tasks are defined as follows:
\begin{itemize}
    \item \textbf{AutoCheckout}: Weigh a product, scan its barcode, and hand a payment QR code to the customer.
    \item \textbf{PackBreads}: Place various differently-shaped bread items into a designated bread box.
    \item \textbf{PackStationery}: Collect three marker pens and a whiteboard eraser, and place them into a packaging box.
    \item \textbf{SortToCabinet}: Stow two distinct objects into separate cabinet drawers, requiring the robot to open and close the drawers twice.
\end{itemize}

\begin{table}[ht]
    \centering
    \caption{$\pi_{0.5}$ Policy Success Rates on Synthesized Tasks}
    \label{tab:vla_training}
    \vskip 0.15in
    \begin{small}
    \begin{sc}
    \resizebox{\columnwidth}{!}{%
    \begin{tabular}{lccc}
        \toprule
        \textbf{Task Name} & \textbf{Pre-train SR} & \textbf{Raw SR} & \textbf{Compressed SR} \\
        \midrule
        AutoCheckout & 0\% & 54\% & 52\% \\
        PackBreads & 0\% & 54\% & 50\% \\
        PackStationery & 0\% & 100\% & 100\% \\
        SortToCabinet & 0\% & 25\% & 28\% \\
        \bottomrule
    \end{tabular}%
    }
    \end{sc}
    \end{small}
    \vskip -0.1in
\end{table}

Table \ref{tab:vla_training} reports the performance of the $\pi_{0.5}$ model, evaluated over 100 trials per task. The \textit{Pre-train SR} column represents the baseline zero-shot success rate prior to fine-tuning. The \textit{Raw SR} and \textit{Compressed SR} columns denote the success rates after supervised fine-tuning on our original uncompressed datasets and our heavily compressed datasets, respectively.

As shown, the $\pi_{0.5}$ model exhibited a 0\% success rate across all tasks prior to training. Following fine-tuning on our datasets, the policy demonstrates a strong acquisition of complex behaviors, achieving a 100\% success rate on \textit{PackStationery} and 54\% on \textit{AutoCheckout}. 

A deeper empirical investigation into the failure cases reveals a strong inverse correlation between the task execution horizon and the final policy success rate. Notably, for tasks with lower success rates, such as \textit{SortToCabinet} (25\%), the majority of failures do not stem from complex articulation sub-tasks, such as opening or closing the cabinet drawers. Instead, the bottleneck predominantly lies in fundamental grasping failures. Because long-horizon tasks are composed of sequential atomic operations, a single missed grasp at any intermediate step cascades into a total task failure. Consequently, the overall success rate becomes exponentially sensitive to the length of the task horizon. Nevertheless, successfully elevating the performance from a 0\% baseline to reliable execution rates across such extended horizons strictly demonstrates the high quality and effectiveness of the V-CAGE training data in instilling robust visuo-motor representations for multi-step reasoning.

\subsection{Sim-to-Real Transfer and Robustness Evaluation}
\label{subsec:sim2real}

\begin{figure}[ht]
    \centering
    \includegraphics[width=\linewidth]{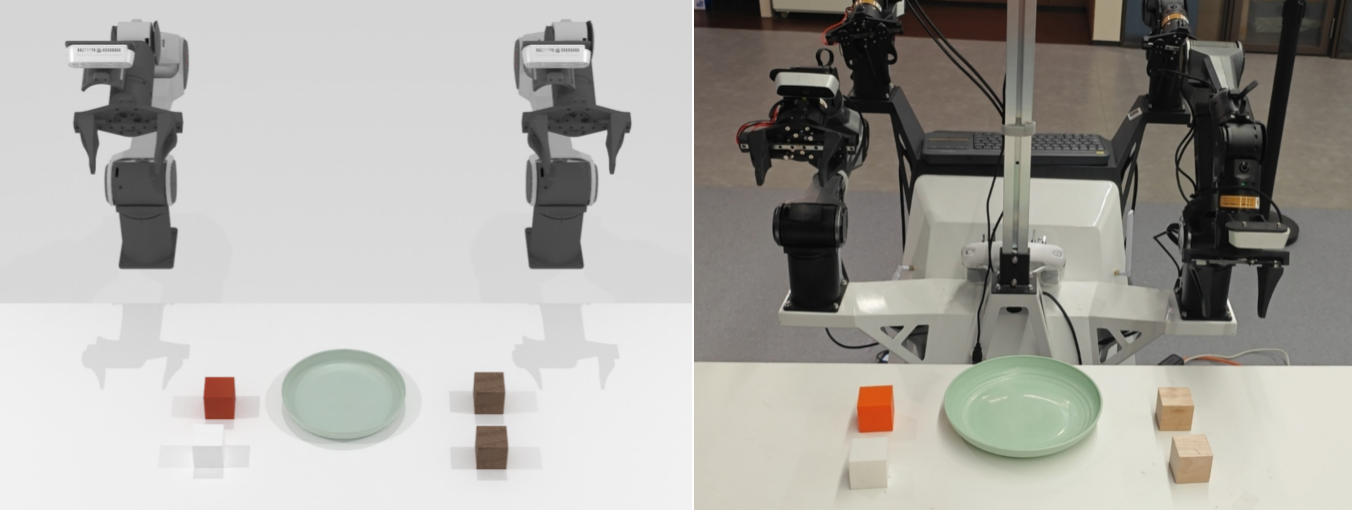}
    \caption{Sim-to-Real evaluation setup on the ALOHA-AgileX platform. The robot is tasked with grasping and placing four blocks into a designated plate.}
    \label{fig:sim2real}
\end{figure}

To evaluate how our synthesized simulation data enhances the robustness of real-world policies, we conducted Sim-to-Real (Sim2Real) transfer experiments using the ALOHA-AgileX hardware platform. The evaluation task requires the robot to sequentially grasp four blocks and place them into a target plate. 

We collected a minimal set of 10 real-world expert demonstrations alongside 250 simulation trajectories generated by our pipeline. To ablate the performance gains, we trained the $\pi_{0.5}$ model under two data regimes: Real-Only (10 real trajectories) and Co-trained (10 real + 250 sim trajectories). Each policy was rigorously evaluated for 20 trials in the physical environment.

\begin{table}[ht]
    \centering
    \caption{Real-World Evaluation Success Rates (20 trials per setting)}
    \label{tab:sim2real}
    \vskip 0.15in
    \begin{small}
    \begin{sc}
    \resizebox{0.9\columnwidth}{!}{%
    \begin{tabular}{lcc}
        \toprule
        \textbf{Training Regime} & \textbf{Dataset Composition} & \textbf{Success Rate} \\
        \midrule
        Real-Only & 10 Real & 20\% \\
        Co-trained & 10 Real + 250 Sim & 55\% \\
        \bottomrule
    \end{tabular}%
    }
    \end{sc}
    \end{small}
    \vskip -0.1in
\end{table}

Table \ref{tab:sim2real} summarizes the real-world deployment results. As expected, training on an extremely constrained real-world dataset (Real-Only) yields a marginal success rate of 20\%, highlighting the data-hungry nature of VLA models. 

Importantly, co-training the model with both domains yields a 55\% success rate. This absolute improvement of 35\% over the baseline directly demonstrates that our simulation data is highly transferable and effectively acts as a robust data-augmentation mechanism, significantly alleviating the burden of real-world data collection while enhancing physical execution stability.

\subsection{Impact of Video Data Compression}
\label{subsec:compression}
Given the substantial storage footprint required for scaling large-scale visual trajectory data, we evaluated whether applying our custom video compression algorithm degrades downstream VLA performance. We retrained the $\pi_{0.5}$ model using heavily compressed datasets for all four tasks. 

As detailed in Table \ref{tab:vla_training}, the policy success rates trained on the compressed data are highly comparable to those trained on the uncompressed raw data. Specifically, the performance differences fall well within the expected variance of deep reinforcement learning and behavioral cloning: \textit{AutoCheckout} (52\% vs. 54\%), \textit{PackBreads} (50\% vs. 54\%), \textit{PackStationery} (100\% vs. 100\%), and \textit{SortToCabinet} (28\% vs. 25\%). This demonstrates that our compression technique effectively preserves the critical spatial and semantic features necessary for complex manipulation, offering a highly efficient solution for large-scale data curation without sacrificing downstream learning efficacy.

\subsection{Ablation Study: Scene Generation and Verification}
\label{subsec:ablation}

\textbf{Effectiveness of the IGSC Module.} We first evaluate the necessity of the Inpainting-Guided Scene Construction (IGSC) module. To quantify its impact, we compare the sub-task generation quality of our pipeline against a baseline employing random object placement. In long-horizon tasks, particularly those involving multi-object interactions, a critical requirement for automated data collection is the logical spatial arrangement of multiple interacting objects. Since our logic for synthesizing long-horizon tasks from sub-tasks follows a Markovian assumption, the success rate of individual sub-tasks directly reflects the variance in the overall success rate of the long-horizon tasks. We evaluated the sub-task success rates generated by our method across various scene categories, as detailed in Table~\ref{tab:igsc_ablation}.

As shown, the IGSC module consistently outperforms the random layout baseline in the proportion of valid tasks (i.e., tasks achieving a success rate $>0\%$) across all tested environments. Notably, in highly constrained spaces like the \textit{Cabinet}, the random baseline completely fails to generate any executable tasks (0\% valid ratio), whereas IGSC unlocks a 66.7\% valid task ratio. Similarly, in complex settings such as the \textit{Office Desk} and \textit{Kitchen}, IGSC significantly boosts the valid task proportion from approximately 55\% to over 72\%. Although the mean success rate of valid tasks fluctuates slightly across certain categories (e.g., \textit{Office Desk}) due to the inherent complexity of the newly generated spatial layouts, the substantial increase in the volume of executable sub-tasks highlights that IGSC effectively eliminates geometric conflicts and enforces contextual priors. This marked improvement underscores the necessity of semantically coherent scene layouts for scalable data synthesis.

\begin{table}[ht]
    \centering
    \caption{Ablation of IGSC on Sub-Task Generation Across Scenes} 
    \label{tab:igsc_ablation}
    \vskip 0.15in
    \begin{small}
    \resizebox{\columnwidth}{!}{%
    \begin{sc}
    \setlength{\tabcolsep}{3pt} 
    \begin{tabular}{llccc}
        \toprule
        \textbf{Scene} & \textbf{Tasks} & \textbf{Metric} & \textbf{Random} & \textbf{IGSC (Ours)} \\
        \midrule
        \multirow{2}{*}{Bathroom} & \multirow{2}{*}{9}
        & Valid ($>0\%$) & 55.6\% & \textbf{66.7\%} \\
        & & Mean SR & 80.0\% & 80.0\% \\
        \midrule
        \multirow{2}{*}{Cabinet} & \multirow{2}{*}{9}
        & Valid ($>0\%$) & 0.0\% & \textbf{66.7\%} \\
        & & Mean SR & 0.0\% & \textbf{33.3\%} \\
        \midrule
        \multirow{2}{*}{Cashier Desk} & \multirow{2}{*}{11}
        & Valid ($>0\%$) & 54.5\% & \textbf{72.7\%} \\
        & & Mean SR & 93.3\% & \textbf{95.0\%} \\
        \midrule
        \multirow{2}{*}{Kitchen} & \multirow{2}{*}{33}
        & Valid ($>0\%$) & 54.5\% & \textbf{72.7\%} \\
        & & Mean SR & 73.3\% & \textbf{88.3\%} \\
        \midrule
        \multirow{2}{*}{Office Desk} & \multirow{2}{*}{43}
        & Valid ($>0\%$) & 55.8\% & \textbf{81.4\%} \\
        & & Mean SR & 87.5\% & 85.1\% \\
        \bottomrule
    \end{tabular}
    \end{sc}%
    }
    \end{small}
    \vskip -0.1in
\end{table}

\textbf{Impact of VLM-Based Closed-Loop Verification.} Next, we ablate the VLM-based closed-loop verification module to quantify its role in mitigating dataset noise. We compare the full framework against an open-loop generation baseline across four complex, long-horizon tasks. In the open-loop setting (w/o VLM), the system executes programmatic data generation without visual feedback. This approach inevitably suffers from ``silent failures''---instances where the script completes without system exceptions, yet the physical task semantics (e.g., maintaining a stable grasp or correctly placing an object) fail. The closed-loop setting (w/ VLM) employs our VLM module as a rigorous visual critic to evaluate and filter these generated trajectories. Table~\ref{tab:vlm_ablation} reports the validity of the synthesized data batches under both configurations.

The empirical results highlight the strict necessity of the verification module. Without the VLM, the generated datasets contain significant execution noise. The actual success rate of the unverified data (Ground Truth Success Ratio) fluctuates between 45.45\% and 81.82\%, meaning a large portion of the generated trajectories are false positives that would severely degrade downstream policy learning. By acting as a rigorous visual critic, our VLM effectively performs rejection sampling. The integration of this module achieves a 100\% precision rate across all four tasks, meaning every trajectory approved by the VLM is physically and semantically correct. This strictly validates the module's critical function in pruning execution noise and ensuring the high fidelity of the synthesized training data.

\begin{table}[ht]
    \centering
    \caption{Impact of VLM Verification on Data Generation Purity}
    \label{tab:vlm_ablation}
    \vskip 0.15in
    \begin{small}
    \resizebox{\columnwidth}{!}{%
    \begin{sc}
    \setlength{\tabcolsep}{6pt} 
    \begin{tabular}{lcc}
        \toprule
        \textbf{Task Name} & \textbf{Open-Loop (w/o VLM)} & \textbf{V-CAGE (w/ VLM)} \\
        \midrule
        AutoCheckout & 63.64\% & \textbf{100.00\%} \\
        PackBreads   & 81.82\% & \textbf{100.00\%} \\
        PACKSTATIONERY &100\% & \textbf{100.00\%} \\
        SortToCabinet & 72.72\% & \textbf{100.00\%} \\
        PickBlocks   & 45.45\% & \textbf{100.00\%} \\
        \bottomrule
    \end{tabular}
    \end{sc}%
    }
    \end{small}
    \vskip -0.1in
\end{table}

\section{Conclusion}
In this work, we presented V-CAGE, a closed-loop framework for synthesizing high-fidelity, long-horizon manipulation datasets. 
By shifting from scripted automation to an agentic architecture, V-CAGE orchestrates a complete synthesis pipeline—from task-oriented scene configuration to verified trajectory collection. Our Inpainting-Guided Scene Construction (IGSC) algorithm ensures the physical validity of complex environments by eliminating geometric conflicts, while the integration of the OpenClaw agent enables the autonomous execution of diverse, high-level semantic plans. We use VLM-based closed-loop verification mechanism serves as a rigorous filter, "severing" the error propagation chain and ensuring that only causally consistent trajectories are retained. Furthermore, our action-aware perceptual compression resolves the "storage wall" by achieving over 90\% filesize reduction without compromising visual fidelity for downstream VLA training. 
Experimental results demonstrate that V-CAGE-generated data significantly enhances policy robustness to cluttered environments, enabling generalization to unseen tasks. These findings demonstrate that the combination of physically-consistent scene synthesis and autonomous verification is essential to producing high-quality data required for robust embodied intelligence.


\textbf{Limitations and Future Work.} While V-CAGE ensures high data fidelity, the rejection sampling process can be computationally expensive for highly complex tasks with low inherent success probabilities. Future work could explore using the VLM critic to provide dense reward signals for correcting failed trajectories in-the-loop, rather than discarding them, thereby improving generation efficiency. Additionally, extending V-CAGE to dynamic interactions involving deformable objects or fluids remains an exciting direction for future research.






\bibliographystyle{IEEEtran}
\bibliography{conference}

\vspace{12pt}

\end{document}